\documentclass[11pt,a4paper]{article}
\usepackage[hyperref]{acl2018}
\usepackage{times}
\usepackage{latexsym}
\usepackage{url}
\usepackage{amsmath,graphicx,caption,color,epsfig,float,pbox,tabularx,wrapfig,mathrsfs,multirow,amsfonts,amsthm,times,subfigure,amssymb,mathrsfs,algorithm}

\usepackage{graphicx}
\usepackage{multirow}
\usepackage{amsmath}
\usepackage{array}
\usepackage{tabulary}
\newcolumntype{K}[1]{>{\centering\arraybackslash}p{#1}}

\aclfinalcopy % Uncomment this line for the final submission
\def\aclpaperid{66} %  Enter the acl Paper ID here

\title{Concept Transfer Learning for Adaptive Language Understanding}

\author{Su Zhu \and Kai Yu  \thanks{\ \ The corresponding author is Kai Yu.}\\
  Key Laboratory of Shanghai Education Commission for Intelligent Interaction and Cognitive Engineering\\
  SpeechLab, Department of Computer Science and Engineering\\
  Brain Science and  Technology Research Center\\
  Shanghai Jiao Tong University, Shanghai, China\\
  {\tt \{paul2204,kai.yu\}@sjtu.edu.cn} \\}

\date{}

\begin{document}
\maketitle
\begin{abstract}
Concept definition is important in language understanding (LU) adaptation since literal definition difference can easily lead to data sparsity even if different data sets are actually semantically correlated. To address this issue, in this paper, a novel concept transfer learning approach is proposed. Here, substructures within literal concept definition are investigated to reveal the relationship between concepts. A hierarchical semantic representation for concepts is proposed, where a semantic slot is represented as a composition of {\em atomic concepts}. Based on this new hierarchical representation, transfer learning approaches are developed for adaptive LU. The approaches are applied to two tasks: value set mismatch and domain adaptation, and evaluated on two LU benchmarks: ATIS and DSTC 2\&3. 
Thorough empirical studies validate both the efficiency and effectiveness of the proposed method. In particular, we achieve state-of-the-art performance ($F_1$-score 96.08\%) on ATIS by only using lexicon features.
\end{abstract}

\section{Introduction}
\label{sec:intro}

The language understanding (LU) module is a key component of dialogue system (DS), parsing user's utterances into corresponding semantic concepts (or semantic slots \footnote{Slot and concept are equal in LU. They will be mixed in the rest of this paper to some extent.}). For example, the utterance \emph{``Show me flights from Boston to New York"} can be parsed into \emph{(from\_city=Boston, to\_city=New York)} \cite{pieraccini1992speech}. Typically, the LU is seen as a plain slot filling task. With sufficient in-domain data and deep learning models (e.g. recurrent neural networks, bidirectional long-short term memory network), statistical methods have achieved satisfactory  performance in the slot filling task recently \cite{kurata-EtAl:2016:EMNLP2016, vu2016sequential, liu2016attention}.

However, retrieving sufficient in-domain data for training LU model \cite{tur2010left} is unrealistic, especially when the semantic slot extends or dialogue domain changes. The ability of LU approaches to cope with changed domains and limited data is a key to the deployment of commercial dialogue systems (e.g.  Apple Siri, Amazon Alexa, Google Home, Microsoft Cortana etc).

\begin{figure}[]
\centering
\includegraphics[width=7cm,height=4.5cm]{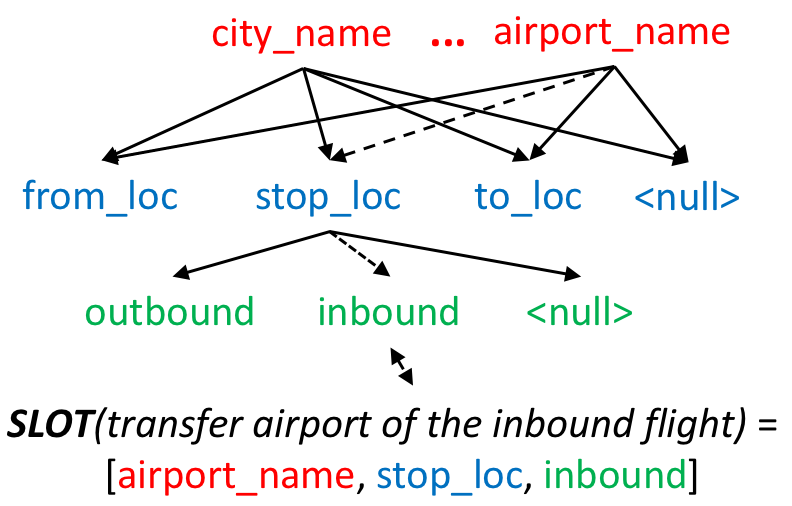}
\caption{An example of hierarchical structure to represent semantic slot with atomic concepts. There are three levels in this structure. The plain slot $SLOT(\text{\emph{transfer airport of the inbound flight}})$ can be represented as a tuple of atomic concepts sequentially.}
\label{fig:tree}
\end{figure}

In this paper, we investigate substructure of semantic slots to find out slot relations and promote data reuse. We represent semantic slots with a hierarchical structure based on atomic concept tuple, as shown in Figure \ref{fig:tree}. Each semantic slot is composed of different atomic concepts, e.g. slot \emph{``from\_city"} can be defined as a tuple of atoms $[\emph{``from\_location"}, \emph{``city\_name"}]$, and \emph{``date\_of\_birth"} can be defined as $[\emph{``date"}, \emph{``birth"}]$.

%Atomic concepts can be classified into two categories, one is value-aware (domain-independent) and the other is context-aware (domain-specific) (e.g. \emph{``city\_name"} and \emph{``date"} are value-aware, \emph{``stop\_loc"}, \emph{``inbound"}  and \emph{``birth"} are context-aware).  Modelling on the atomic concepts but not the plain slot helps find out the linguistic patterns of related slots by semantic sharing, and even decrease the required amount of data. 

\begin{figure}[]
\centering
\includegraphics[width=7.7cm,height=1.3cm]{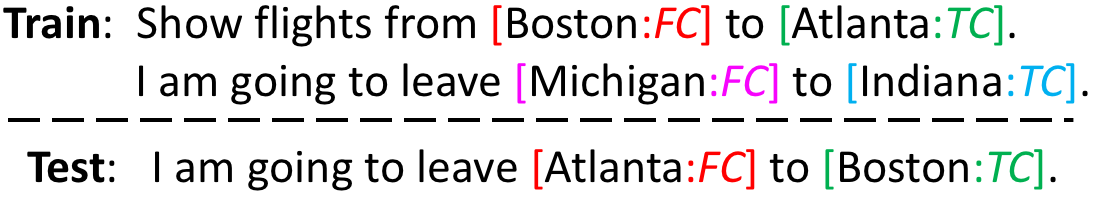}
\caption{An example of mismatched LU datasets labelled with \emph{[value: slot]}. \emph{FC} refers to \emph{``from\_city"}. \emph{TC} refers to \emph{``to\_city"}. }
\label{fig:utt}
\end{figure}

Unlike the traditional slot definition on a plain level, modeling on the atomic concepts helps identify linguistic patterns of related slots by atom sharing, and even decrease the required amount of training data. For example, the training and test sets are unmatched in Figure \ref{fig:utt}, whereas the patterns of atomic concepts (e.g. \emph{``from"}, \emph{``to"}, \emph{``city"}) can be shared.

%We also propose learning a slot filling model on the granularity of atomic concepts by considering these atoms independently or dependently. 
In this paper, we investigate the slot filling task switching from plain slots to hierarchical structures by proposing the novel atomic concept tuples which are constructed manually. For comparison, we also introduce a competitive method which automatically learns slot representation from the word sequence of each slot name. Our methods are applied to value set mismatch and domain adaptation problems on ATIS \cite{Hemphill1995The} and DSTC 2\&3 \cite{henderson2013dialog} respectively. As shown in the experimental results, the slot-filling based on concept transfer learning is effective in solving the value set mismatch and domain adaptation problems. The concept transfer learning method especially achieves state-of-the-art performance ($F_1$-score 96.08\%) on the ATIS task.

%Our main contributions can be summarized as:
%\begin{itemize}
%\item We investigate slot filling task switching from plain slots to hierarchical structures by proposing the novel atomic concept tuples.
%\item The slot-filling based on concept transfer learning shows a good way to solve the value set mismatch and domain adaptation problems, as shown in the experimental results. The concept transfer learning method especially achieves the state-of-the-art performance ($F_1$-score 96.08\%) in the ATIS task. %, only using the lexicon features. 
%\end{itemize}

The rest of the paper is organized as follows. The next section is about the relation to prior work. The atomic concept tuple is introduced in section \ref{sec:atomic}. The proposed concept transfer learning is then described in section \ref{sec:transferlearning}. Section \ref{sec:slotname} describes a competitive method with slot embedding derived from the literal descriptions of slot names. In section \ref{sec:exp}, the proposed approach is evaluated on the value set mismatch and domain adaptation problems. Finally, our conclusions are presented in section \ref{sec:conclusion}.

\section{Related Work}
\label{sec:relatedwork}

\textbf{Slot Filling in LU} \citet{Zettlemoyer2007Online} proposed a grammar induction method by learning a Probabilistic Combinatory Categorial Grammar (PCCG) from logical-form annotations. As a grammar-based method, PCCG is close to a hierarchical concepts structure in grammar generation and combination. But this grammar-based method does not possess high generalization capability for atomic concept sharing, and heavily depends on a well-defined lexicon set.

Recent research on statistical slot filling in LU has been focused on the Recurrent Neural Network (RNN) and its extensions. At first, RNN outperformed CRF (Conditional Random Field) on the ATIS dataset \cite{yao2013recurrent, mesnil2013investigation}. Long-short term memory network (LSTM) was introduced to obtain a marginal improvement over RNN \cite{yao2014spoken}. After that, many RNN variations were proposed: encoder-labeler model \cite{kurata-EtAl:2016:EMNLP2016}, attention model \cite{liu2016attention, zhu2016encoder} etc. However, these work only predicted the plain semantic slot, not the structure of atomic concepts.

\textbf{Domain Adaptation in LU} For the domain adaptation in LU, \citet{Zhu2014Semantic} proposed generating spoken language surface forms by using patterns of the source domain and the ontology of the target domain. With regard to the unsupervised LU, \citet{Heck2012Exploiting} exploited the structure of semantic knowledge graphs from the web to create natural language surface forms of entity-relation-entity portions of knowledge graphs. For the zero-shot learning of LU, \citet{Ferreira2015Zero, Yazdani2015A} proposed a model to calculate similarity scores between an input sentence and semantic items. In this paper, we focus on the extension of slots with limited seed data.

\section{Atomic Concept Tuples}
\label{sec:atomic}

Although concept definition is one of the most crucial problems of LU, there is no unified surface form for the domain ontology. Even for the same semantic slot, names of this slot may be quite different. For example, the city where the flight departs may be called  \emph{``from\_city"}, \emph{``depart\_city"} or \emph{``from\_loc.city\_name"}. Ontology definitions from different groups may be similar but not consistent, which is not convenient for data reuse. Meanwhile, semantic slots defined in traditional LU systems are on a plain level, while there is no structure to indicate their relation.

To solve this problem, we propose to use atomic concepts to represent the semantic slots. Atomic concepts are exploited to break down the slots. We represent the semantic slots as atomic concept tuples (Figure \ref{fig:tree} is an example). The semantic slot composed of these atomic concepts can keep a unified resource for concept definition and extend the semantic knowledge flexibly.

We propose a criteria to construct atomic concept manually. For a given vocabulary $C$ of the atomic concepts, a semantic slot $s$ can be represented by a tuple $[c_1, c_2, ..., c_k]$, where $c_i \in C$ is in the $i$-th dimension and $k$ is tuple length. In particular, a  \emph{``null"} atom is introduced for each dimension.  Table \ref{concept_branch} illustrates an example of slot representation on the ATIS task. To avoid a scratch concept branch, we make a constraint:
\begin{equation*}
\begin{split}
C_i \cap C_j &= \{null\}, 1 \leq i \neq j \leq k 
\end{split}
\end{equation*}
where $C_i$ ($1 \leq i \leq k$) denotes all possible atomic concepts which exist in dimension $i$ (i.e. $c_i \in C_i$). The concept tuple is ordered.  %For example, if we define two slots [\emph{``city\_name", ``fromloc"}] and [\emph{``state\_name", ``toloc"}], then the city where the flight arrives should be defined as [\emph{``city\_name", ``toloc"}] but not [\emph{``toloc", ``city\_name"}]. 

In general, atomic concepts can be classified into two categories, one is value-aware and the other is context-aware. The principle for defining slot as a concept branch is: lower dimension less context-aware. For example, \emph{``city\_name"} and  \emph{``airport\_name"} depend on rare context (value-aware). They should be located in the first dimension. \emph{``from\_location"} depends on the context like a pattern of  \emph{``a flight leaves [city\_name]"}, which should be in the second dimension. The atomic concept tuple shows the inner relation between different semantic slots explicitly. 

\begin{table}[h]
\begin{center}
\begin{tabular}{|l|l|}
\hline \bf slot & \bf atomic concept tuple \\ \hline
city &  [\emph{city\_name}, \emph{null}] \\
from\_city &   [\emph{city\_name}, \emph{from\_location}] \\
depart\_city &   [\emph{city\_name}, \emph{from\_location}] \\
arrive\_airport &   [\emph{airport\_name}, \emph{to\_location}] \\
\hline
\end{tabular}
\end{center}
\caption{\label{concept_branch} An example of slot representation by atomic concepts.}% in the ATIS task. }
\end{table}

Therefore, the procedure of constructing atomic concept tuples for slots can be divided into the following steps.
\begin{itemize}
\item Firstly, we build a vocabulary $C$ of the atomic concepts for all the slots. By analyzing the conceptual intersection of different slots, we can split the slots into smaller ones which are called atomic concepts. After that, each slot is represented as a set of atomic concepts which are not ordered.
\item Secondly, we gather the atoms into different groups. Atomic concepts from the same group should be mutually exclusive. Therefore we can investigate the inner relation and outer relation of these groups.
\item Finally, each group is associated with one dimension ($C_i$) of the atomic concept tuple. The groups are ordered depending on whether they are value-aware or context-aware.
\end{itemize}

%The handcrafted atomic concepts are used in this paper. In future work, we will try to extract atomic concepts automatically. \tf{remove this sentence?}

\section{Concept Transfer Learning}
\label{sec:transferlearning}

\begin{figure}[htbp]
\centering
\includegraphics[width=7.7cm,height=0.8cm]{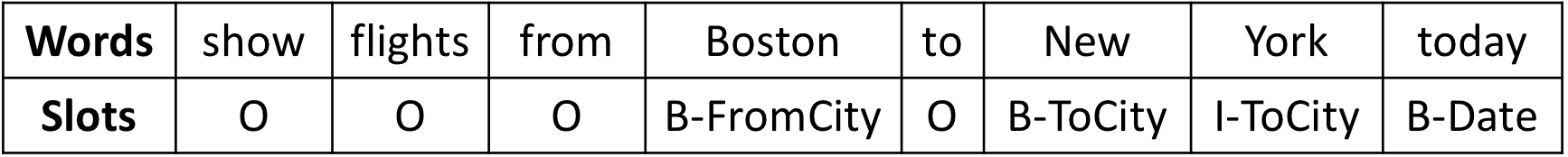}
\caption{An example of annotation for slot filling.}
\label{fig:atis}
\end{figure}

The slot filling is typically considered as a sequence labelling problem. In this paper, we only consider the sequence-labelling based slot filling task. The input (word) sequence is denoted by $\textbf{w} = (w_1, w_2, ..., w_N)$, and the output (slot tag) sequence is denoted by $\textbf{s} = (s_1, s_2, ..., s_N)$. Since a slot may be mapped to several continuous words, we follow the popular in/out/begin (IOB) representation (e.g. an example in Figure \ref{fig:atis}).

\begin{figure*}[]
\centering
\includegraphics[width=13cm,height=5.3cm]{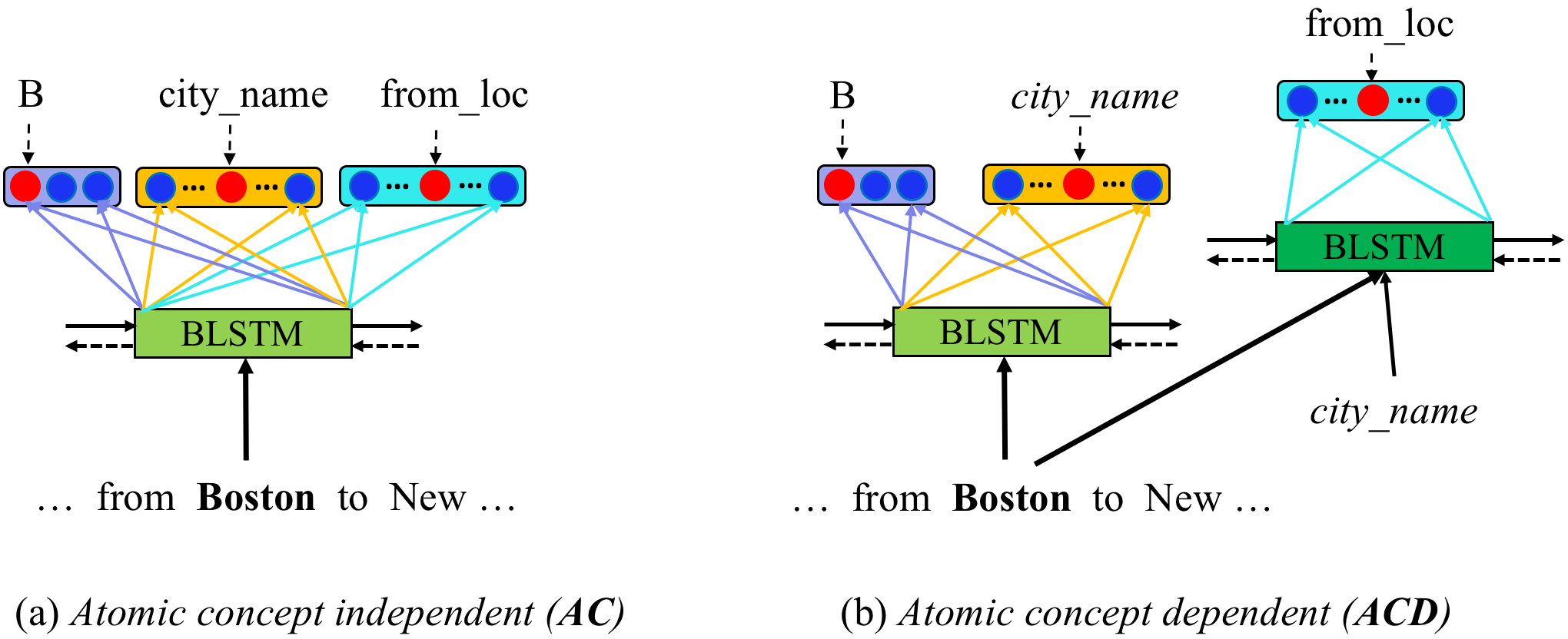}
%\caption{Examples for atomic-concepts based slot-filling in the ATIS task.}
\caption{The proposed method about the atomic-concepts based slot filling. A slot is considered as a tuple of atomic concepts, e.g. \emph{``from\_city"} is represented as $[\emph{``city\_name"}, \emph{``from\_loc"}]$. Multiple output layers are utilized to predict different atoms (including IOB schema). We involve two architectures: a) the AC assumes that the output layers are independent, b) while the ACD makes a dependence assumption.}
\label{fig:abc}
\end{figure*}

The typical slot filling task predicts a plain slot sequence given a word sequence, dubbed as \textbf{plain slot-filling} (\textbf{PS}). %, since the slot is regarded as a single and united symbol, like \emph{``B-fromloc.city\_name"} in Figure \ref{fig:abc}(a). 

In this paper, the popular bidirectional LSTM-RNN (BLSTM) is used to model the sequence labeling problem ~\cite{Graves2012Supervised}. It can be exploited to capture both past and future features for a specific time frame. 
The BLSTM reads the input sentence $\textbf{w}$ and generates $N$ hidden states $h_i=\overleftarrow{h_{i}}\oplus\overrightarrow{h_{i}}, i \in \{1,..,N\}$: 
\begin{equation*}
\overleftarrow{h_{i}}=b(\overleftarrow{h_{{i+1}}}, e_{w_i});\ \overrightarrow{h_{i}}=f(\overrightarrow{h_{{i-1}}}, e_{w_i})
\end{equation*}
where $\overleftarrow{h_{i}}$ is the hidden vector of the backward pass in BLSTM and $\overrightarrow{h_{i}}$ is the hidden vector of the forward pass in BLSTM at time $i$, $b$ and $f$ are LSTM units of the backward and forward passes respectively, $e_w$ denotes the word embedding for each word $w$, and $\oplus$ denotes the vector concatenation operation. We write the entire operation as a mapping $\text{BLSTM}_{\Theta^w}$ ($\Theta^w$ refers to the parameters):
\begin{equation}
(h_1 ... h_N) = \text{BLSTM}_{\Theta^w}(w_1 ... w_N)
\label{eqn:blstm}
\end{equation}

Therefore, the plain slot filling defines a distribution over slot tag sequences given an input word sequence: 
\begin{equation}
\begin{split}
p(\textbf{s}|\textbf{w}) &= \prod_{i=1}^N p(s_i|h_i)\\
&= \prod_{i=1}^N \text{softmax}(W_o \cdot h_i)^T \delta_{s_i}
\end{split}
\label{eqn:sf}
\end{equation}
where the matrix $W_o$ (output layer) consists of the vector representations of each slot tag, the symbol $\delta_{d}$ is a Kronecker delta with a dimension for each slot tag, and the \emph{softmax} function is used to estimate the probability distribution over all possible plain slots.

\subsection{Atomic-Concepts Based Slot Filling}

The slot is indicated as an atomic concept tuple based on hierarchical concept structure. Slot filling is considered as a concept-tuple labelling task. 

\textbf{(a) Atomic concept independent}

Slot filling can be transferred to a multi-task sequence labelling problem, regarding these atomic concepts \textbf{independently} (i.e. \textbf{AC}). Each task predicts one atomic concept by a respective output layer. Thus, the slot filling problem can be formulated as
\begin{equation*}
p(\textbf{s}|\textbf{w}) = \prod_{i=1}^N [p(\text{IOB}_i|h_i) \prod_{j=1}^k p(c_{ij}|h_i)]
\end{equation*}
where the semantic slot $s_i$ is represented by an atomic concept branch $[c_{i1}, c_{i2}, ..., c_{ik}]$, and $\text{IOB}_i$ is the IOB schema tag at time $i$. As illustrated in Figure \ref{fig:abc}(a), the semantic slot ``\emph{from\_city}" can be represented as $[\emph{``city\_name"}, \emph{``from\_loc"}]$. The prediction of IOB is regarded as another task specifically. All tasks share the same parameters except for the output layers.

\textbf{(b) Atomic concept dependent}

Atomic concepts can also be regarded \textbf{dependently} (i.e. \textbf{ACD}) so that atomic concept prediction depends on the former predicted results. The slot filling problem can be formulated as
\begin{equation*}
\begin{aligned}
&p(\textbf{s}|\textbf{w}) \\
=& \prod_{i=1}^N [p(\text{IOB}_i|h_i) p(c_{i1}|h_i) \prod_{j=2}^k p(c_{ij}|h_i, c_{i,1:j-1})]
\end{aligned}
\end{equation*}
where $c_{i,1:j-1} = (c_{i,1}, ..., c_{i,j-1})$ is the predicted result of former atomic concepts of slot tag $s_i$, indicating a structured multi-task learning framework.

In this paper, we make some simplifications on concept dependence. We predict atomic concept only based on the last atomic concept, as shown in Figure \ref{fig:abc}(b). 

%Actually we simplify the tree structure into hierarchical structure in modelling for preliminary investigation. Therefore, the model may produce an arbitrary concept branch undefined in the atomic concept trees.

\subsection{Training and Decoding}

Since our approach is a structured multi-task learning problem, the model loss is summed over each task during training. For the domain adaptation, we firstly gather training data from the source domain and seed data from the target domain to be a union set. Subsequently, the union data is fed into the slot filling model.

During the decoding stage, we combine predicted atomic concepts with probability multiplication. The evaluation is made on the top-best hypothesis. Although the atomic-concepts based slot filling may predict an unseen slot. We didn't perform any post-processing but considered the unseen slot as a wrong prediction.

\section{Literal Description of Slot Name}
\label{sec:slotname}

In the section, we introduce a competitive system which uses the literal description of the slot as an input of the slot filling model. The literal description of slot used in this paper is the word sequence of each slot name, which can be obtained automatically. As the names of relative slots may include the same or similar word, the word sequence of slot name can also help reveal the relation between different slots. Therefore, it is very meaningful to compare this method with the atomic concept tuples involving human knowledge.

\begin{figure}[htbp]
\centering
\includegraphics[width=7.7cm,height=4.5cm]{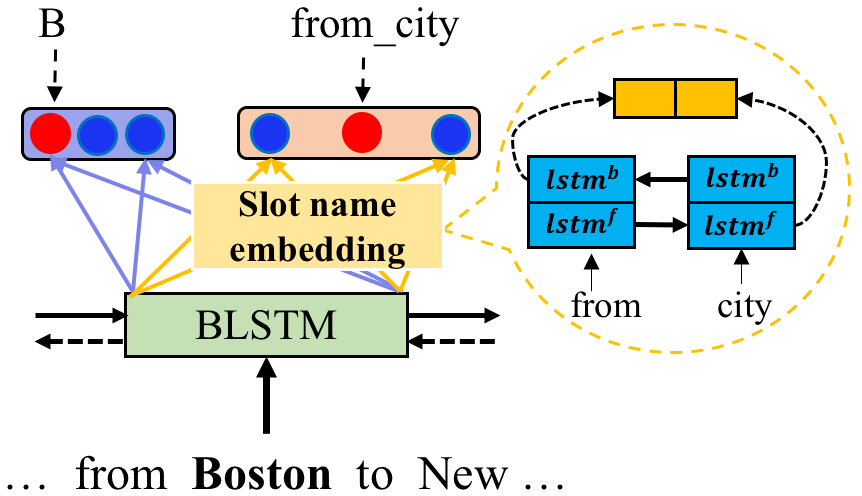}
\caption{The proposed framework of slot filling based on the literal description of the slot. The literal description of a slot is the word sequence of slot name which can be obtained automatically, e.g. \emph{``from\_city"} is represented as a word sequence of ``from city". Another BLSTM in the orange dotted circle is exploited to derive softmax embeddings from the slot names.}
\label{fig:smm}
\end{figure}

The architecture of this competitive system is illustrated in Figure \ref{fig:smm}. First, it assumes that each slot name is a meaningful natural language description so that the slot filling task is tractable from the input word sequence and slot name. Second, another BLSTM model is applied to derive softmax embedding from the slot names. In this method, we also split the slot filling task into IOB tag prediction and slot name prediction. In other words, the slot tag $s_i$ is broken down into $\text{IOB}_i$ and slot name $\text{SN}_i$,  e.g. the slot tag ``B-\emph{from\_city}" is split into ``B" and ``\emph{from\_city}". The details are indicated below.

With the BLSTM applied on the input sequence, we have hidden vectors $h_i, i \in \{1,..,N\}$ as shown in Eqn. (\ref{eqn:blstm}). This model redefines the distribution over slot tag sequences given an input word sequence, compared with Eqn. (\ref{eqn:sf}): 
\begin{equation*}
p(\textbf{s}|\textbf{w}) = \prod_{i=1}^N p(\text{IOB}_i|h_i)p(\text{SN}_i|h_i)
\end{equation*}
where $p(\text{IOB}_i|h_i)$ predicts the IOB tag and $p(\text{SN}_i|h_i)$ makes a prediction for the slot name. We define
\begin{equation*}
p(\text{SN}_i|h_i) = \text{softmax}(W \cdot h_i)^T \delta_{\text{SN}_i}
\end{equation*}
where $W \in \mathbb{R}^{A\times B}$ is a matrix, $h_i \in \mathbb{R}^{B}$ is a vector, $A$ is the number of all different slot names. The matrix $W$ consists of the embedding of each slot name (i.e. each row vector of $W$ with length $B$).

To capture the slot relation within different slot names,  we apply another BLSTM model (as shown in the orange dotted circle of Figure \ref{fig:smm}) onto the word sequence (literal description) of each slot name. For the $j$-th slot name ($j \in \{1,..,A\}$) with a word sequence $\textbf{x}^j = (x_1^j, ..., x_{N_j}^j)$, we have
\begin{equation*}
\overleftarrow{v_{n}^j}=\text{lstm}^b(\overleftarrow{v_{{n+1}}^j}, e_{x_n^j});\ \overrightarrow{v_{n}^j}=\text{lstm}^f(\overrightarrow{v_{{n-1}}^j}, e_{x_n^j}) 
\end{equation*}
where $\overleftarrow{v_{n}^j}$ is the hidden vector of the backward pass and $\overrightarrow{v_{n}^j}$ is the hidden vector of the forward pass at time $n$ ($n \in \{1,..,N_j\}$), $e_x$ denotes the word embedding for each word $x$. We take the tails of both backward and forward pass as the slot embedding, i.e.
\begin{equation*}
W_j = \overleftarrow{v_{1}^j} \oplus \overrightarrow{v_{N_j}^j}
\end{equation*}
where $W_j$ is the $j$-th row vector of matrix $W$.
%where $\oplus$ denotes the vector concatenation operation.

The relative slots using the same or similar word in slot naming will be close in the space of slot embedding inherently. Therefore, this method is a competitive system to the atomic concept tuples. We will show the comparison in the following section.

\section{Experiments}
\label{sec:exp}

We evaluate our atomic-concept methods on two tasks: value set mismatch and domain adaptation. 

\textbf{Value set mismatch} task evaluates the generalization capability of different slot filling models. In a language understanding (LU) system, each slot has a value set with all possible values which can be assigned to it. Since the semantically annotated data is always limited, only a part of values is seen in the training data. Will the slot filling model perform well on the unseen values? To answer this question, we synthesize a test set by the values mismatched with the training set of ATIS corpus. Our methods may take advantages of the prior knowledge about slot relations based on the atomic concepts and the literal descriptions of slot names.

\textbf{Domain adaptation} task evaluates the adaptation capability of our methods when they meet new slots in the target domain. In this task, a seed training set of the target domain is provided. However, it is very limited: 1) some new slots may not be covered; 2) not all contexts are covered for each new slot. The atomic-concepts based method would alleviate this problem. Each slot is defined as a tuple of atomic concepts in our method. Therefore, it is possible to learn an unseen slot of the target domain if its atomic concepts exist in the data of the source domain and the seed data of the target domain. It is also possible to see more contexts for a new slot if its atomic concepts exist in the source domain which has much more data.

\subsection{Value Set Mismatch}

\textbf{ATIS} corpus has been widely used as a benchmark by the LU community. The training data consists of 4978 sentences and the test data consists of 893 sentences. 

In this task, we perform an adaptation for unmatched training and test sets, in which there are many unseen slot-value pairs in the test set (Figure \ref{fig:utt} is an example). It is a common problem in the development of commercial dialogue system since it is impossible to collect data covering all possible slot-value pairs. We simulate this problem on the ATIS dataset \cite{Hemphill1995The} by creating an unmatched test set (\textbf{ATIS\_X\_test}).

ATIS\_X\_test is synthesized from the standard ATIS test set by randomly replacing the value of each slot with an unseen one. The unseen value sets are collected from the training set according to bottom-level concepts (e.g. \emph{``city\_name"}, \emph{``airport\_name"}). For example, if the value set of \emph{``from\_city"} is \{\emph{``New York"}, \emph{``Boston"}\} and the value set of \emph{``to\_city"} is \{\emph{``Boston"}\}, then the unseen value for \emph{``to\_city"} is \emph{``New York"}.  The test sentence \emph{``Flights to [xx:to\_city]"}  can be replaced to \emph{``Flights to [New York:to\_city]"}. Finally, the ATIS\_X\_test gets the same sentence number to the standard ATIS test set.

\subsubsection{Experimental Settings}

We randomly selected 80\% of the training data for model training and the remaining 20\% for validation. We deal with unseen words in the test set by marking any words with only one single occurrence in the training set as $\langle  unk \rangle$. We also converted sequences of numbers to the string DIGIT, e.g. “1990” is converted to “DIGIT*4” \cite{ZhangA}. Regarding BLSTM model, we set the dimension of word embeddings to 100 and the number of hidden units to 100. For training, the network parameters are randomly initialized in accordance with the uniform distribution (-0.2, 0.2). Stochastic gradient descent (SGD) is used for updating parameters. The \emph{dropout} with a probability of 0.5 is applied to the non-recurrent connections during the training stage.
%Only the current word is used as input without any context words.

We try different learning rates by grid-search in range of $[0.008, 0.04]$. We keep the learning rate for 100 epochs and save the parameters that give the best performance on the validation set. Finally, we report the $F_1$-score of the semantic slots on the test set with parameters that have achieved the best $F_1$-score on the validation set. The $F_1$-score is calculated using CoNLL evaluation script. \footnote{http://www.cnts.ua.ac.be/conll2000/chunking/output.html}

\subsubsection{Experimental Results and Analysis}

%We compare our systems with the published results on the standard ATIS test set. The results are shown in Table \ref{tab:compare}. 

Table \ref{tab:compare} summarizes the recently published results on the ATIS slot filling task and compares them with the results of our proposed methods on the standard ATIS test set.  We can see that RNN outperforms CRF because of the ability to capture long-term dependencies. LSTM beats RNN by solving the problem of vanishing or exploding gradients. BLSTM further improves the result by considering both the past and future features. Encoder-decoder achieves the state-of-the-art performance by modeling the label dependencies. Encoder-labeler is a similar method to the Encoder-decoder. These systems are designed to predict the plain semantic slots traditionally. 

Compared with the published results, our method outperforms the previously published F1-score, illustrated in Table \ref{tab:compare}. \textbf{AC} gets a marginal improvement (+0.15\%) over \textbf{PS} by predicting the atomic concepts independently instead of the plain slots. Moreover, \textbf{ACD} predicts the atomic concepts dependently, gains 0.50\% (significant level 95\%) over the \textbf{AC}. Worth to mention that \textbf{ACD} achieves a new state-of-the-art performance of the standard slot-tagging task on the ATIS dataset, with only the lexicon features \footnote{There are other published results that achieved better performance by using Name Entity features, e.g. \citet{mesnil2013investigation} got 96.24\% $F_1$-score. The NE features are manually annotated and strong information. So it would be more meaningful to use only lexicon features. Meanwhile, several other works can obtain competitive results by using the intent classification as another task for joint training, e.g. \citet{liu2016attention} achieved 95.98\% $F_1$-score. In this paper, we consider the slot filling task only.}. 

%To the best of our knowledge, we are the first to report a $F_1$-score over 96\% using only the lexicon features.

Our methods are also tested on the ATIS\_X\_test to measure the ability of generalization. For comparison, we also apply dictionary features (n-gram indication) of value sets (e.g. some kind of gazetteers) collected from training data into the \textbf{PS} model (i.e. PS+\emph{dict-feats} in Table \ref{tab:compare}). From Table \ref{tab:compare}, we can see that: 1) The plain slot filling models (\textbf{PS}, Encoder-decoder) are not on par with other models. 2) The atomic-concepts based slot filling gets a slight improvement over the \textbf{PS} with \emph{dict-feats}, considering the concepts independently (\textbf{AC}). 3) The atomic-concepts based slot fillings (\textbf{ACD} gains a large margin over \textbf{AC}, considering the concepts dependently. 4) The method based on slot name embedding (described in Section \ref{sec:slotname}) achieves a slight improvement than \textbf{AC}, which implies that it is possible to reveal the relationship between slots automatically. 

\begin{table} []
\centerline{
%\small
\begin{tabular}{p{4cm}K{0.9cm}K{1.4cm}}
\hline
Model & {\small ATIS} & {\small ATIS\_X\_test} \\
\hline  \hline
CRF \cite{mesnil2013investigation} & 92.94 & --   \\
RNN \cite{mesnil2013investigation} & 94.11 & --   \\
LSTM \cite{yao2014spoken} & 94.85 & --    \\
BLSTM \cite{ZhangA} & 95.14 & --  \\
Encoder-decoder \cite{liu2016attention}  & 95.72 & --  \\
Encoder-labeler \cite{kurata-EtAl:2016:EMNLP2016}  & 95.66 & --  \\
Encoder-decoder-pointer \cite{zhai2017neural} & 95.86 & --  \\
\hline \hline
Encoder-decoder$^*$  & 95.79 & 79.84   \\
BLSTM$^*$ (PS)  & 95.43 & 79.59 \\
PS + dict-feats & 95.57 & 80.74 \\
AC & 95.58 & 80.90  \\
ACD & \bf{96.08} & \bf{86.16}  \\
Slot name embedding & 95.52 & 81.49 \\
\hline
\end{tabular}
}
\caption{\label{tab:compare} {Comparison with the published results on the standard ATIS task, and evaluation on ATIS\_X\_test. ($*$ denotes our implementation.)}}
\end{table}

\begin{table*}
\begin{center}
\centerline{
%\small
\begin{tabular}{l|l}
\hline
\bf Reference &  ... could get in [boston:city\_name]  [late:\textbf{period\_of\_day}]  [night:period\_of\_day] \\ \hline
\bf PS & ... could get in [boston:city\_name]  [late:\textbf{airport\_name}]  [night:period\_of\_day] \\
\bf AC & ... could get in [boston:city\_name]  [late:\textbf{period\_of\_day}]  [night:period\_of\_day] \\
\bf ACD & ... could get in [boston:city\_name]  [late:\textbf{period\_of\_day}]  [night:period\_of\_day] \\
\hline
\end{tabular}
}
\end{center}
\caption{\label{tab:atis_case} Examples show how concept transfer learning benefits. We use \emph{[value:slot]} for annotation.}
\end{table*}

\textbf{Case study}: As illustrated in Table \ref{tab:atis_case}, the plain slot filling (PS) predicts the label of \emph{``late"} wrongly, whereas the atomic-concepts based slot fillings (i.e. AC and ACD) get the accurate annotation. The word of ``late" is never covered by the slot \emph{``period\_of\_day"} in the training set. It is hard for the plain slot filling (PS) to predict an unseen mapping correctly. Luckily, the ``late" is covered by the family of the slot \emph{``period\_of\_day"}  in the training set, e.g. \emph{``arrive\_time.period\_of\_day"}. Therefore, AC and ACD can learn this by modeling the atomic concepts separately.

\subsection{Domain Adaptation}

%\subsubsection{Datasets}

Our methods are also evaluated on the DSTC 2\&3 task \cite{henderson2013dialog} which is considered to be a realistic domain adaptation problem. 

\textbf{DSTC 2 (source domain)} comprises of dialogues from the restaurant information domain in Cambridge. We use the \emph{\bf dstc2\_train} set (1612 dialogues) for training and the \emph{\bf dstc2\_dev} (506 dialogues) for validation.

\textbf{DSTC 3 (target domain)} introduces the tourist information domain about restaurant, pubs and coffee shops in Cambridge,  which is an extension of DSTC 2. We use seed data \emph{\bf dstc3\_seed} (only 11 dialogues) as the training set of the target domain.

\textbf{DSTC3\_S\_test}: In this paper, we focus on three new semantic slots: \emph{``has\_tv, has\_internet, children\_allowed"}. \footnote{For each slot of \emph{``has\_tv, has\_internet, children\_allowed"}, the semantic annotation \emph{``request(slot)"} is replaced with \emph{``confirm(slot=True)"}. Then we have the slot-tagging format, e.g. "does it have [television:confirm.has\_tv]".} They only exist in the DSTC 3 dataset and have few appearances in the seed data. A test set is chosen for specific evaluation on these new semantic slots, by gathering all the sentences (688 sentences) whose annotation contains these three slots and randomly selecting 1000 sentences irrelevant to these three slots from the \emph{dstc3\_test} set. This test set is named as \emph{\bf DSTC3\_S\_test} (1688 sentences).

The union of a slot and action is taken as a plain semantic slot (e.g. \emph{``confirm.food=Chinese"}), since each slot is tied with an action (e.g. \emph{``inform"}, \emph{``deny"} and \emph{``confirm"}) in DSTC 2\&3. The slot and action are taken as atomic concepts. For the slot filling task, only the semantic annotation with aligned information is kept, e.g. the semantic tuple \emph{``request(phone)"} is ignored. We use transcripts as input, and make slot-value alignment by string matching simply. %spoken value matching simply.

\subsubsection{Experimental Results and Analysis}

The experimental settings are similar to the ATIS's, whereas the seed data in DSTC 3 is also used for validation.

\begin{table} [htbp!]
\centerline{
\begin{tabular}{ccc}
\hline
Model & Training set & $F_1$-score  \\
\hline  \hline
PS & dstc3\_seed & 83.52     \\
PS & dstc2\_train + dstc3\_seed & 89.57 \\
\hline
AC & dstc3\_seed & 83.58  \\
AC  & dstc2\_train + dstc3\_seed & 91.98\\
ACD & dstc2\_train + dstc3\_seed & \textbf{92.15}    \\
\hline
\end{tabular}
}
\caption{\label{tab:compare_dstc} {The performance of our methods evaluated on the DSTC3\_S\_test.}} % AC(D) denotes atomic concept (dependent).}}
\end{table}

The performance of our methods in the DSTC 2\&3 task is illustrated in Table \ref{tab:compare_dstc}. We can see that: 1) By incorporating the data of the source domain (dstc2\_train), \textbf{PS} and \textbf{AC} achieve improvements respectively. 2) \textbf{AC} gains more than \textbf{PS} by modeling the plain semantic slot as atomic concepts. The atomic concepts promote the associated slots to share input features for the same atoms. 3) The atomic-concepts based slot filling considering the concepts dependently (\textbf{ACD}) gains little (0.17\%) over \textbf{AC} considering the concepts independently. It may be due to the small size of dstc3\_seed.

\textbf{Case study}: Several cases from these models (trained on the union set of dstc2\_train and dstc3\_seed) are also chosen to explain why the atomic-concepts based slot filling outperforms the typical plain slot filling, as shown in Table \ref{tab:case_dstc}. From the above part of Table \ref{tab:case_dstc}, we can see \textbf{PS} predicts a wrong slot. Because the grammar \emph{``does it have [something]"} is only for the plain slot \emph{``confirm.hastv"} in the seed data. From the below part of Table \ref{tab:case_dstc}, we can see that only \textbf{ACD} which considers the concepts dependently predicts the right slot. Since \emph{``confirm.childrenallowed"} never exists in the seed data, \textbf{PS} can't learn patterns about it. Limited by the quantity of the seed data, \textbf{AC} also doesn't extract the semantics correctly.

\begin{table} [htbp!]
\vspace{2mm}
\centerline{
\small
\begin{tabular}{|l|l|}
\hline
\bf Reference & does it have [internet:confirm.hasinternet] \\ \hline
\bf PS  & does it have [internet:confirm.hastv] \\
\bf AC & does it have [internet:confirm.hasinternet] \\
\bf ACD & does it have [internet:confirm.hasinternet] \\
\hline  \hline
\bf Reference & do they allow [children:confirm.CA] \\ \hline
\bf PS  & do they allow [children:CA] \\
\bf AC & do they allow [children:CA] \\
\bf ACD & do they allow [children:confirm.CA] \\
\hline
\end{tabular}
}
\caption{\label{tab:case_dstc} {Examples show how concept transfer learning benefits. CA denotes \emph{childrenallowed}.}}
\end{table}

\section{Conclusion} %and Future Work}
\label{sec:conclusion}

To address data sparsity problem of language understanding (LU) task, we present a novel method of concept definition based on well-defined atomic concepts. We present the concept transfer learning for slot filling on the atomic concept level to solve the problem of adaptive LU. The experiments on the ATIS and DSTC 2\&3 datasets show our method obtains promising results and outperforms the traditional slot filling, due to the knowledge sharing of atomic concepts. 

The atomic concepts are constructed manually in this paper. In future work, we want to explore more flexible concept definition for concept transfer learning of LU. Moreover, we also propose a competitive method based on slot name embedding which can be extracted from the literal description of the slot name automatically. The experimental result shows that it lays foundation for finding a more flexible concept definition method for adaptive LU.

\section*{Acknowledgments}

%This work has been supported by the National Key Research and Development Program of China under Grant No.2017YFB1002102, and the China NSFC projects (No. 61573241). 

This work has been supported by the China NSFC project (No. 61573241), Shanghai International Science and Technology Cooperation Fund (No. 16550720300) and the JiangSu NSFC project (BE2016078). Experiments have been carried out on the PI supercomputer at Shanghai Jiao Tong University.  We also thank Tianfan Fu for comments that greatly improved the manuscript.

%The acknowledgments should go immediately before the references.  Do not number the acknowledgments section ({\em i.e.}, use \verb|\section*| instead of \verb|\section|). Do not include this section when submitting your paper for review.

%\bibliography{acl2018}
\bibliography{acl2018}
\bibliographystyle{acl_natbib}

\end{document}